# An innovative mixed reality approach for Robotics Surgery

**Gabriela Rus [1], Nadim Al Hajjar [2], Ionut Zima[1], Calin Vaida[1], Corina Radu[3], Damien Chablat [1,4], Andra Ciocan[2] and Doina Pîslă[1,5*]**

[1] CESTER, Research Center for Industrial Robots Simulation and Testing, Technical University of Cluj-Napoca, 400641 Cluj-Napoca, Romania
[2] Department of Surgery, "Iuliu Hatieganu" University of Medicine and Pharmacy, 400347 Cluj-Napoca, Romania
[3] Department of Internal Medicine, "Iuliu Hatieganu" University of Medicine and Pharmacy, 400347 Cluj-Napoca, Romania
[4] Nantes Université, École Centrale Nantes, CNRS, LS2N, UMR 6004, F-44000 Nantes, France
[5] Technical Sciences Academy of Romania, 26 Dacia Blvd, 030167 Bucharest, Romania

*E-mail: doina.pisla@mep.utcluj.ro

**Abstract.** Robotic-assisted procedures offer numerous advantages over traditional approaches, including improved dexterity, reduced fatigue, minimized trauma, and superior outcomes. However, the main challenge of these systems remains the poor visualization and perception of the surgical field. The goal of this paper is to provide an innovative approach concerning an application able to improve the surgical procedures offering assistance in both preplanning and intraoperative steps of the surgery. The system has been designed to offer a better understanding of the patient through techniques that provide medical images visualization, 3D anatomical structures perception and robotic planning. The application was designed to be intuitive and user friendly, providing an augmented reality experience through the Hololens 2 device. It was tested in laboratory conditions, yielding positive results.

## 1. Introduction

Procedures involving critical organs like the pancreas and esophagus have advanced significantly in the field of medical robots. Robotic technologies have transformed surgical procedures on these essential organs, giving surgeons greater control and precision as they navigate complex anatomical structures. Robotic platforms provide greater dexterity for surgeries involving the pancreas, which are sometimes complex due to the organ's location and function. Similarly, the accuracy and minimally invasiveness of robotic-assisted techniques have revolutionized surgeries involving the esophagus, with its intricate network of tissues and close proximity to important structures. Consequently, patients undergoing esophageal and pancreatic procedures had better postoperative results, shorter recovery periods, and less stress, highlighting the critical role that medical robotics plays in these types of procedures. [1-2].

The favourable results provided by assisted robotic surgery have led to widespread acceptance of robots in these procedures, with recent decades witnessing accelerated development of many structures designed for this purpose [3-5]. These structures have proven particularly suitable for procedures targeting organs located in hard-to-access areas such as the pancreas or oesophagus. However, despite the undeniable advantages, robots used for these types of procedures bring with them some limitations, such as appropriate visualization of the surgical field. In order to compensate for this limitation, devices have begun to be used to provide insights about the patient during the procedure, augmenting the surgeon's environment.

Within the domain of medical robotics Augmented Reality (AR) techniques find applications in rehabilitation, surgery, and medical assistance. Within surgical contexts, AR is integrated across various stages, from preoperative planning [6] to intraoperative guidance [7] and telesurgery [8].

In [9] and [10], the authors investigated the application of an AR interface to optimize trocar placement, reducing the risk of robot arm collisions in laparoscopic procedures and facilitating robot configuration initialization in teleoperated robot-assisted surgeries with the da Vinci system.

Fu et al. [11] developed an AR-assisted robotic learning framework tailored for MIS tasks, utilizing an AR interface through the HoloLens 2 OST-HMD to validate the configuration of a serial redundant robot. To establish a reliable trajectory for replication by the robotic system in minimally invasive surgery scenarios, the Gaussian mixture model (GMM) and Gaussian mixture regression (GMR) were employed to encode multiple trajectories demonstrated by humans in this study.

Despite advancements in AR applications for surgery, there's a noticeable gap in comprehensive solutions covering both preplanning and intraoperative phases. Furthermore, just few applications address simulations and the integration of robotic systems. Bridging this gap requires collaborative efforts to design AR solutions benefiting surgical teams throughout the entire procedure.

The aim of the paper is to present an application designed to improve surgical procedures by offering insights into the patient during both preplanning and intraoperative phases. The proposed system is designed to: (i) provide an adequate overview of robotic systems integration in the operating room, assisting both surgeons and engineers in finding suitable solutions for RAS, (ii) facilitate access to medical images and anatomical details of the patient during surgery to improve real-time guidance, and (iii) offer a superior method for pre-planning through features such as 3D anatomical structure visualization, the ability to highlight abnormalities, and visualization of volume rendering to anticipate potential challenges. Following the introduction in section 1, section 2 provides the methodology for developing the application, including a description of the structural architecture. Section 3 presents the results, covering both the design and performance of the application. Finally, section 4 encapsulates the conclusions drawn from this study.

## 2. Methodology

*2.1 Description of the architectural structure*
The application has been developed in Unity, using the MRTK (Mixed Reality Toolkit) packages which provides a set of tools and components used for developing mixed reality applications across a variety of platforms, primarily focusing on Microsoft HoloLens and Windows Mixed Reality devices.

The inputs for the application, including CT scans, 3D organ reconstructions, and the CAD models of the robotic structures, have been provided by other software tools such as Slicer 3D.

Volume data has been exported in Nearly Raw Raster Data (NRRD) format, while 3D reconstructions are in OBJ format. Additionally, Siemens NX was used to export the robotic structures in STEP format, as illustrated in Figure 1.

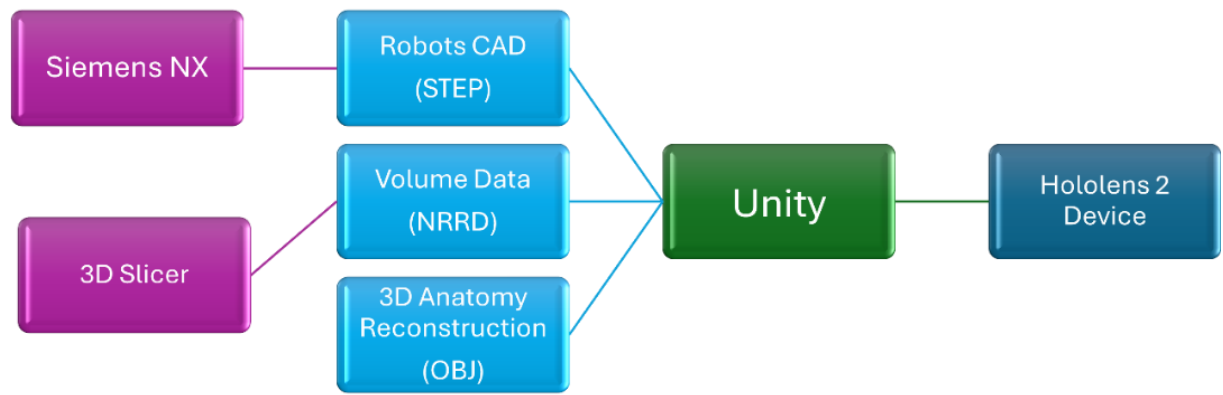

**Figure 1.** Architectural structure of the proposed application

*2.2 Description of the targeted procedures*

For a better understanding of the importance of this application in surgery, an overview of the complex task imposed by pancreatic and oesophagus surgery is required.

### A) Minimally invasive techniques for esophagectomy

Esophagectomy, essential for the treatment of malignant and benign diseases, relies on surgical team's expertise and tumor location. Minimally invasive techniques, including robotics, offer advantages but pose challenges such as high risk of anastomotic leaks. The ongoing Robotic-Assisted Minimally Invasive Esophagectomy (RAMIE) trial evaluates robotic-assisted procedures' efficacy, showing promising early results. These procedures can be approached through three methods.

1. **Transhiatal esophagectomy**

This procedure is specific for tumors in the upper or middle oesophagus.

The procedure is performed as follows: the patient is positioned supine with a steep reverse Trendelenburg tilt, and trocars are placed for robotic assistance and liver retraction. The robot is positioned for optimal tumor visualization. Dissection begins along the greater curvature of the stomach, followed by division of the pars flaccida and creation of a window in the lesser curvature. The left gastric artery is isolated and divided, and the oesophagus is mobilized transhiatally all the way to the cervical region. A cervicectomy is performed. After examining perfusion with ICG, the gastric conduit is anastomosed to the specimen, and the robot is undocked.

2. **Robotic Ivor Lewis esophagectomy**

The patient is positioned supine with a reverse Trendelenburg tilt. Trocars are placed for robotic assistance, assistant support, and liver retraction. Ligaments are divided, arteries and veins are prepared, and an omental flap is created. Mediastinal and nodal dissections is performed. The gastric conduit is created, mobilized in the thorax, where the anastomosis will be done. A jejunostomy tube is inserted at the end and incisions are closed.

3. **Robotic McKeown esophagectomy**

The patient is positioned in a left lateral decubitus, semi-prone, with a 45° anterior tilt. Trocars are placed for robotic assistance and the assistant. The right lung is deflated and retracted after dividing the inferior pulmonary ligament, and the oesophagus is mobilized up after the azygos cross is stapled. Periesophageal and mediastinal lymphadenectomy is performed. The oesophagus is dissected and resected, and the specimen is retrieved. The abdominal time implies greater and lesser curvature of the stomach dissection. With further mobilization of the gastric conduct in the cervical region, where the anastomosis will be done. After closure of incisions, the robot is undocked [12].

### B) Minimally invasive techniques for pancreatectomy

Robotic surgery is increasingly utilized for pancreatic lesions and malignancies providing advantages such reduced post-operative pain, precise manoeuvrability (allows for cutting and suturing at angles not possible for humans' hands) safety and feasibility [13]. The most standard procedures are:

#### 1. Cephalic pancreatoduodenectomy (Whipple procedure)

The procedure, performed with robotic assistance involves positioning the patient supine and inserting trocars for camera and robotic arms. The surgeon mobilizes and transects the head of the pancreas, entire duodenum, gallbladder with the common bile duct, with or without pylorus preservation, meticulously preserving vital surrounding structures such as the portal and superior mesenteric vein and artery. Lymphadenectomy is mandatory around the liver hilum and celiac trunk. Then, reconstruction time involves anastomoses to restore gastrointestinal continuity and pancreatic function: Wirsungo-jejunostomy or pancreato-gastrostomy, hepatico-jejunostomy and gastro-jejunostomy [14].

#### 2. Distal pancreatectomy with or without spleen preservation

The patient is positioned supine, and trocars are inserted for camera and robotic arms. The surgeon dissects and resects the distal pancreas, with peripancreatic lymphadenectomy with spleen preservation or en-bloc with splenectomy ensuring no iatrogenic trauma of adjacent blood vessels and surrounding organs. Proximal pancreatic stump is carefully stapled to avoid pancreatic leakage and fistula.

*2.2 Robotics systems for minimally invasive surgery*

For the simulator module of the current application, the UR collaborative robot and a new spherical robot developed for MIS procedures were used.

The UR5 has 6 degrees-of freedom (DOF), providing a high degree of flexibility and manoeuvrability [15]. The robot's collaborative features enable it to work alongside surgeons, providing steady support and executing predefined trajectories with high repeatability. This collaboration enhances surgical accuracy while minimizing the risk of tissue damage [16].

The spherical robot developed for MIS is presented in Figure 3. The system provides 3 DOF allowing for one translational and one rotational movement used to achieve the desired position under the insertion point, and another translational movement for insertion/extraction of the instrument from the patient's body. Additionally, a counterweight system dynamically counterbalances the instrument load based on gravity compensation, providing structural stiffness. This structure facilitates seamless interchangeability between instruments and the endoscopic camera, alongside a small footprint that allows for easy integration in the operating room [17].

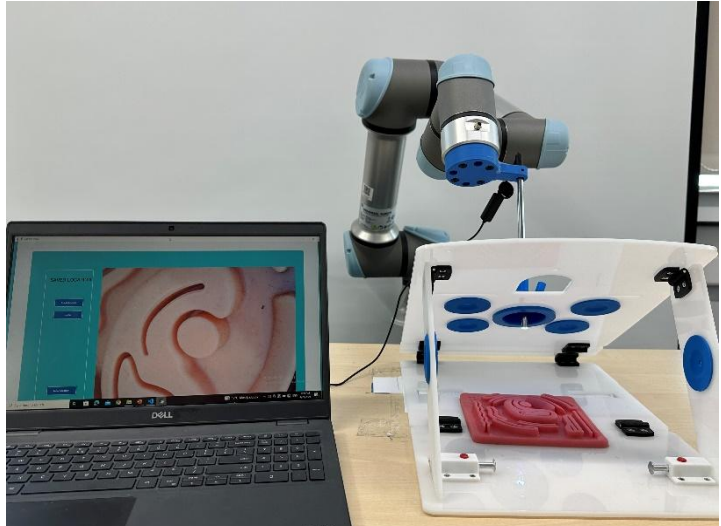

**Figure 2.** The collaborative UR5 robot within the surgical procedure

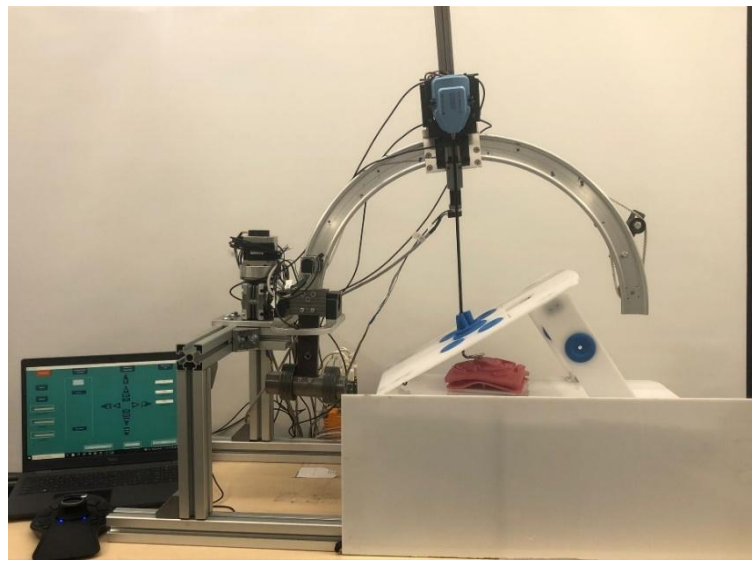

**Figure 3.** Spherical robot within the surgical procedure

*2.3 Volume Data*

CT data, being in Digital Imaging and Communications in Medicine (DICOM) format, has been imported to Slicer 3D program in order to convert the data and to export it in NRRD format, the NRRD offering many advantages such as **simplicity** - NRRD files are typically simpler and more lightweight compared to DICOM files, **ease of Use -** NRRD files are easier to read and manipulate programmatically compared to DICOM files and **reduced overhead** - NRRD files typically have less overhead compared to DICOM files, which can be advantageous in scenarios where storage space or network bandwidth is limited. Considering that Hololens has limitation related to processing and memory, the NRRD format was considered the most suitable. The NRRD file exported was used for both volume rendering and slicers visualizing.

When the NRRD file has been loaded and the raw data in it converted, the next step was to apply the maximum intensity projection (MIP) algorithm. This method is used to calculate the highest intensity value for each viewing ray through a volume, which is then projected onto a 2D display. Consequently, bones or contrast-enhanced regions become more clearly visible while other parts of the image remain unaltered, and this leads to a better final image. After using MIP algorithms on an image volume, voxel values are transformed into colors and opacities through transfer functions that define shades and opacity of materials thus providing distinction between different tissues and improving visual contrast as well. At this point the rendered volume could be viewed as comprehensive 3D representation of all data stored in NRRD file with emphasis on high density structures since these have been facilitated by MIP technique and color-coded basing on transfer function settings.

*2.4 3D Anatomy Reconstruction*
3D Slicer was also used in the 3D reconstruction of anatomical structures. DICOM files were utilized to perform semantic segmentation of the anatomical regions of interest. The segmentation process was automated using the Total Segmentator module provided by the software [18]. Subsequently, the segmented organs were imported into Unity in OBJ format.

*2.5 Hololens 2 device*
HoloLens 2 is designed to merge a virtual world (holograms) with the real environment of the user, creating a mixed reality (MR) experience. This device equipped with a high-definition wave guide display system, coupled with an in-house built holographic processing unit (HPU), for full immersion into visuals and rendering at the speed of thought. The smart glasses are designed to facilitate natural user interaction through improved hand and eye tracking; making it excellent for mapping out 3D environments on which any virtual object could be placed accurately [19].

**3. Results**

*3.1 Functionalities*
The application has developed to have a user-friendly design and intuitive functionality architecture. Thus, the design for this application has been thought to be intuitive (as can be seen in Figure 4), using the MRTK packages which contribute with sliders, buttons, text boxes, etc.

Considering the complexity of the robotic assisted surgery, the application has been developed to offer help both in preplanning and intraoperative phase, providing useful insights about the patient. The functionalities include:

**Slicer visualization** - This functionality allows surgeons to visualize slicers of the CT scans in Hololens 2 device, enhancing spatial understanding and aiding in decision-making. Real-time access to CT scans during surgery enables precise navigation. Furthermore, holographic overlays align virtual structures with patient anatomy, reducing errors. As can be seen in Figure 5 the slicers can be manipulated offering a smooth translation between them.

**Anatomical structure visualization** – functionality enables the medical staff to plan surgical approaches, anticipates potential challenges, and optimizes procedures for better outcomes in preplanning phase and enhance spatial awareness, facilitating precise surgical maneuvers during surgery. As can be seen in Figure 6 this functionality allows to manipulate the anatomical structure of the body separately.

**Volume rendering** – offers comprehensive, contrast-enhanced, and easily interpretable visualizations of anatomical structures, the surgeon being able to identify critical landmarks and navigate complex surgical anatomy. In Figure 7 it can be seen how this is particularly helpful for complex procedures, such as esophagectomy, in which the surgeon needs to remove the tumor using an intercostal approach [20].

**Underlying abnormalities –** can be used for detecting and assessing abnormalities for appropriate diagnosis and treatment planning, as can be seen in Figure 7 a). The application recognizing a gest (tripod grip) and starts draw based on the position of the hand.

**Simulation –** this functionality provides an adequate environment to plan and observe the entire surgical procedure. This component of the application can be very useful especially in robotic assisted surgery, considering the complexity of procedures and the idea that the robots should be integrated in a specific operating room. As demonstrated in Figure 8, the simulation functionality not only offers an environment for planning and observing surgical procedures but also facilitates the visualization of the integration of multiple robots for collaborative medical applications. This feature is particularly valuable in the context of robotic-assisted surgery, where intricate procedures necessitate careful planning and coordination within specialized operating rooms.

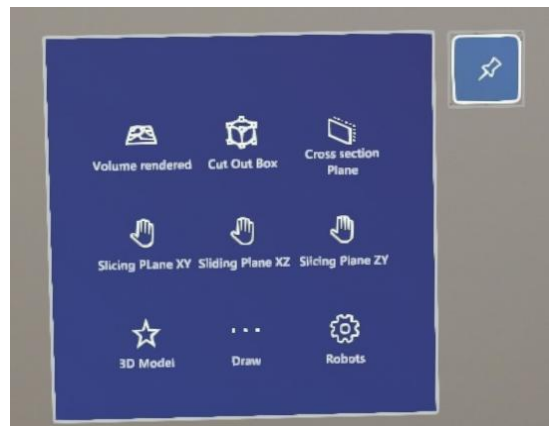

**Figure 4. a)** User Interface visualized from Hololens. The Interface includes buttons for volume rendering, Cut Out Box- direct cut in volume rendering, cross section plane – create a section in the plane, slicers visualization in all 3 planes, visualization of 3D anatomical structures, drawing module, simulations and visualization of surgical robots.

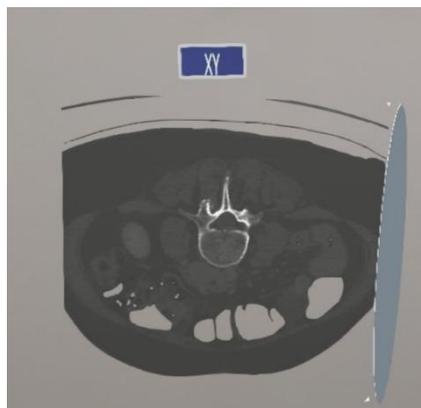 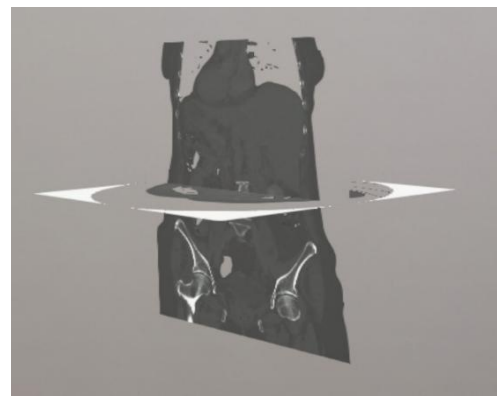

**Figure 5. a)** Slicer visualization – Coronal plane.   **Figure 5. b)** Slicer visualization – Sagittal and Transverse plane

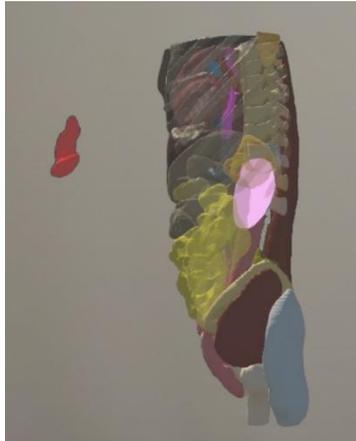

**Figure 6. a)** Anatomical structure visualization – lateral view

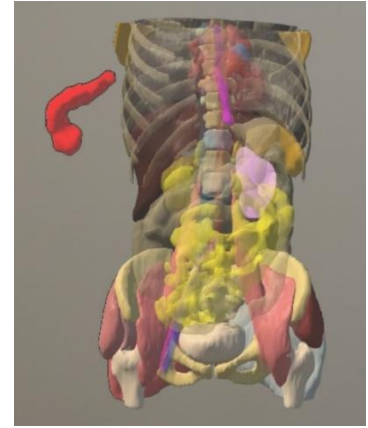

**Figure 6. b)** Anatomical structure visualization – frontal view

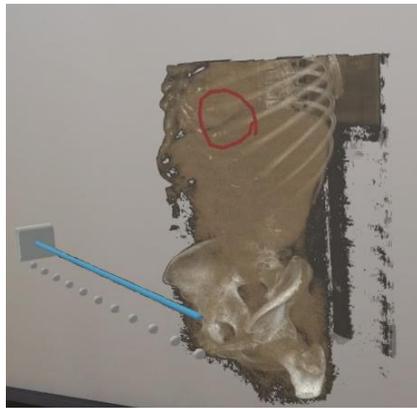

**Figure 7. a)** Visualization of Volume rendering. The functionality provides a slider with which the visible value range can be changed. Within this functionality the surgeon is able to see the desired range

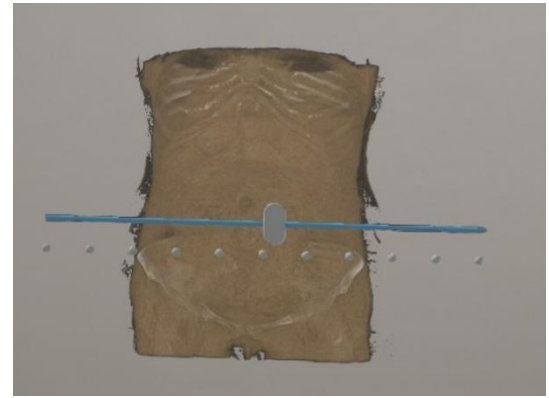

**Figure 7. b)** Visualization of Volume rendering with the ranges of values for the rendering changed

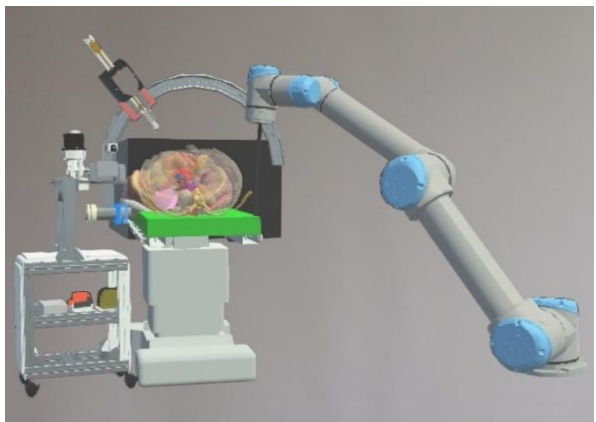

**Figure 8. a)** Simulation module – planning operation with UR5 (as endoscopic holder) and the spherical arm (with the surgical instrument), targeting the pancreas.

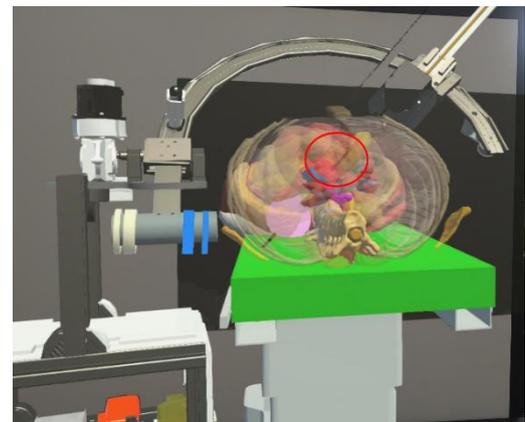

**Figure 8. b)** Simulation module– where the pancreas has been achieved. It can also be observed that the position of the instrument is changed compared to Figure 8.a), with the spherical arm reaching the pancreas.

*3.2 Analysis of the application*

The application was evaluated using a User Experience Questionnaire (UEQ) as a tool. The questionnaire was completed by 10 users in laboratory conditions (not medical procedures), and the results show that the application can be qualified as good, obtaining excellent ratings for Stimulation and Novelty (Figure 9). The main weakness identified is in terms of efficiency, which may be caused by the computational limitations of the HoloLens device [21].

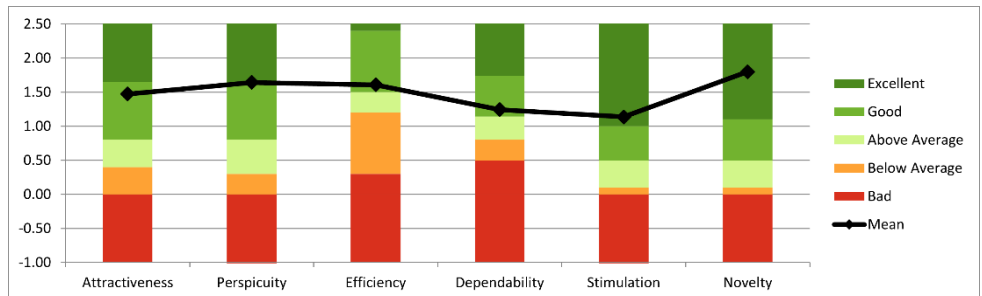

**Figure 9.** Results of UEQ

## 4. Conclusion

This paper focuses on the development and evaluation of an application, based on a mixed reality approach using Hololens 2, designed to enhance minimally invasive surgeries (MIS) by leveraging CT scanner visualization, volume rendering, and robot simulations. The goal of this application is to provide surgeons with valuable information regarding complex anatomical structures, such as the pancreas or oesophagus, thereby enhancing surgical precision and patient outcomes. However, the versatility of this application extends beyond specific anatomical regions, as it can be seamlessly integrated with various robotic platforms, allowing for its utilization across a wide range of surgical procedures. By integrating different robotic systems, surgeons can further customize their approach to each patient's unique needs, ultimately improving the overall efficiency and effectiveness of surgical interventions.

Additionally, the application provides valuable information both about the possibility of integrating robots into the operating room and about the surgery itself, enabling detailed planning. Its collaborative use between medical staff and engineers encourages beneficial cooperation that advances the development of surgical robots.

However, despite the potential benefits offered by the application, its performance is limited by computational weaknesses inherent in the device. It is crucial to consider these constraints in order to maximize the application's effectiveness in MIS procedures. Future research efforts should prioritize the optimization of computational capabilities to ensure seamless operation and user satisfaction. By addressing these challenges, our application has the potential to significantly advance surgical practice and improve patient care in the field of minimally invasive surgery.

**Acknowledgment**

This work was supported by the project New smart and adaptive robotics solutions for personalized minimally invasive surgery in cancer treatment - ATHENA, funded by European Union – NextGenerationEU and Romanian Government, under National Recovery and Resilience Plan for Romania, contract no. 760072/23.05.2023, code CF 116/15.11.2022, through the Romanian Ministry of Research, Innovation and Digitalization, within Component 9, investment I8.